# D-Point Trigonometric Path Planning based on Q-Learning in Uncertain Environments

Ehsan Jeihaninejad, Azam Rabiee

*Abstract*—Finding the optimum path for a robot for moving from start to the goal position through obstacles is still a challenging issue. This paper presents a novel path planning method, named D-point trigonometric, based on Q-learning algorithm for dynamic and uncertain environments, in which all the obstacles and the target are moving. We define a new state, action and reward functions for the Q-learning by which the agent can find the best action in every state to reach the goal in the most appropriate path. The D-point approach minimizes the possible number of states. Moreover, the experiments in Unity3D confirmed the high convergence speed, the high hit rate, as well as the low dependency on environmental parameters of the proposed method compared with an opponent approach.

## I. Introduction

Path planning is the process of finding the shortest feasible path. The planning has been considered as a challenging concern in video games [1], transportation systems [2], and mobile robots [3][4]. As the most important path planning issues, we can refer to the dynamics and the uncertainty of the environment, the smoothness and the length of the path, obstacle avoidance, and the computational cost. In the last few decades, researchers have done numerous research efforts to present new approaches to solve them [5][6][7][8].

Generally, most of the path planning approaches are categorized to one of the following methods [9][10] [11]:
(1) Classical methods
   (a) Computational geometry (CG)
   (b) Probabilistic roadmap (PRM)
   (c) Potential fields method (PFM)
(2) Heuristic and meta heuristic methods
   (a) Soft computing
   (b) Hybrid algorithms

Since the complexity and the execution time of CG methods were high [11], PRMs were proposed to reduce the search space using techniques like milestones [12]. Even though the PRMs cannot present a path with a high quality, but they are much faster than the CGs and easy to implement [12]. On the other hand, most of the methods suffer from the dynamics of the environment [13]. It means that when the environment changes after planning, the agent often requires re-planning, which is not cost-effective as the environment might continually change. When the environment is dynamic and infinitive, applying approaches like cell decomposition are expensive in terms of execution time. Instead, adaptive machine learning approaches, such as the PFM are reliable choices when we have a dynamic and uncertain environment. Li and his colleagues [14] proposed a neural- network-based path planning approach for a multi-robot system in a dynamic environment. They believe that traditional path planning can be expensive in terms of computations. Hence, the neural network-based solutions can be considered as an acceptable alternative.

However, Xin and his colleague [15] proposed a hybrid planning, in which the neural network is used for modelling the environment and the genetic algorithm (GA) is applied for optimizing the Y axis' values of a direct line between the start and the goal point. The GA-based approaches have the potential for solving the path planning problems, as GA is originally an optimization algorithm. However, they suffer from the trade-off between the speed and the accuracy, as well as not properly defining the initial population [12]. Davoodi and his colleagues [16] also attempted to improve path planning using GAs. Proposing two multi-objective optimization approaches, they could improve metrics, such as safety and energy consumption (which are known as the common goal of path planning), smoothness, clearness, and the length of the path. Three later mentioned metrics are known as natural-looking path challenge introduced in [17].

Although the mentioned machine learning methods are capable of planning with an acceptable quality, they suffer from the "local minima" problem [16][18][19][9]. In addition, most of them needs tedious supervised training. As a semi-supervised technique, reinforcement learning (RL) are the best matching algorithms for the path planning [20]. Various kinds of RL-based approaches including Q-learning are reviewed in [19]. Konar and his colleagues [21] proposed an improved version of Q-learning (IQL) for path planning in a deterministic maze environment. For a dynamic environment, Ngai and Yung [22] have proposed double action Q-learning for obstacle avoidance. The authors believe that in a dynamic environment the only action taker is not the agent but also the environment should be

E. Jeihaninejad is with the Young Researchers and Elite Club, Dolatabad Branch, Islamic Azad University, Isfahan, Iran. He got his master degree from the university (e-mail: jeihani.ehsan@gmail.com).

A. Rabiee is assistant professor at Department of Computer Science, Dolatabad Branch, Islamic Azad University, Isfahan, Iran (*corresponding author*, email: a.rabiee@iauda.ac.ir). She is also researcher at KI4AI, KAIST, South Korea.

considered as another action taker. One of the main differences of this methodology with the classical Q-learning (CQL) [23] is the possibility of changing the current state by both the environment and the agent (observers). However, the combination of the double action Q-learning with the CQL have successfully applied for automated vehicle overtaking problem [24]. Furthermore, Kareem and his colleagues [19] used Q-learning algorithm for path planning in a dynamic environment. To define the state space, they utilized a trigonometric method with association of environment segmentation, which is based on the mobile robot location. They believed this state space definition is based on human reasoning model for the path planning. Beside their benefits, the main disadvantages of the Q-learning-based methods are the high number of states, the low convergence speed, and not effective performance. In fact, the larger the number of states gets, the more complex solving the problem becomes.

Inspiring from [19], this paper proposes a novel Q-learning-based trigonometric approach for state space definition called "D-point" to create the minimum possible number of states, for decreasing the convergence time, maximizing the number of successful paths (named hit rate), and reducing the dependence of a successful plan on environmental parameters.

The rest of the paper is organized as fallows. Section II describes the proposed methodology. In the Section III, we analyse our approach and illustrate the experimental results of comparing the opponent and the proposed approaches. Finally, Section IV summarizes the paper.

## II. PROPOSED D-POINT TRIGONOMETRIC PATH PLANNING

Before going through describing our proposed methodology, the assumptions of this study are described and then we review the Q-learning algorithm in the next sub-section. Later, the definition of the 'state space' and the 'action space' of the proposed method are described in the following sub-sections. The 'number of states' and the 'immediate reward function' are also discussed at the end.

### A. Assumptions

1) There is always an attraction that makes the agent goes in a direct path toward the target when it is in the areas which are considered safe.
2) The agent does not have the ability for walking backwards. If necessary to go back, it should do right or left actions twice with 90 degrees rotation.
3) The agent does not have any information about the obstacles' locations, except the nearest one. The agent continuously checks a circle-shaped area around itself to find whether there is any obstacle nearby. However, if there are more than one obstacle in this area it considers the nearest one.
4) In every instance of the time, the agent knows where the target is located.
5) The environment is dynamic and every object in the environment is moving, even the target.
6) The center of the world is the location of the nearest obstacle and is independent from the rotation of the obstacle. In other words, it depends on the x, y and z axis.

### B. Q-Learning algorithm

Generally, machine learning algorithms are divided into two main categories, supervised and unsupervised or semi-supervised learning algorithms. While the former, e.g. the typical artificial neural network, takes advantage of a collection of data for learning, the latter tries to achieve an optimal policy (solution) by autonomously exploring and experiencing predefined possible behaviours in the problem space. Coined by [20], the Q-learning is an unsupervised learning algorithm which is suitable for finding an optimal action-selection policy for any given finite Markov decision process. As the basis of our study, we review a typical Q-learning algorithm for a simple deterministic environment. Suppose that the environment can be define by a finite number of states, $s$. However, there are finite number of possible actions $a$ in every state. The goal is to adapt Q values for every state and action, $Q(s, a)$, such that the $Q$ value shows the suitability of the action $a$ in state $s$. The algorithm can be summarized as follows:

1. For each $s$ and $a$ initialize table entry $Q(s, a)$ to zero.
2. Observe the current state $s_t$
3. Do forever:
    - Select an action $a_t$ and execute it.
    - Receive immediate reward $r(s_t, a_t)$ from the environment
    - Observe the new state $s_{t+1}$
    - Update the table entry $Q(s_t, a_t)$ as
    $$Q(s_t, a_t) = r(s_t, a_t) + \gamma . MAX(Q, s_{t+1}) \qquad (1)$$
    - $s_{t+1} \leftarrow s_t$

Where, in Eq. (1), $r(s_t, a_t)$ is the given immediate reward for taking the action which leads to transferring to the next state $s_{t+1}$. Moreover, $\gamma$ is the discount factor. Finally, $MAX(Q, s_{t+1})$ is the maximum Q-value of the all possible actions in the next state $s_{t+1}$.

### C. States Space Definition

Our objective is to introduce an effective Q-learning-based approach for path planning in a dynamic environment. To achieve this, the number of states should be very low. In the rest of this section, we describe how we achieve our objective.

First, we divide the world into four regions called R1, R2, R3, and R4. As can be seen in Fig. 1, these divisions are based on the center of the world i.e., the location of the nearest obstacle, depicted by a circle. It means when the nearest obstacle changes, the location of the center of the world changes too. Then, for defining the states, we use the regions in which the target and the agent are located. The target and the agent are shown by cross and drop signs, respectively. Moreover, in the figure, the large circle around the nearest obstacle shows the non-safe area.

According to Fig. 1, the current state is define by $s = [R1, R4]$ in which R1 and R4 are the target and agent regions, respectively. In simulations, we obtain the regions of the target and the agent using the differences between their x and y location with respect to the nearest obstacle position. Finally, we need a certain discriminator that makes the states unique from each other. For this purpose, we introduce the D-point approach, which is based on the following linear and trigonometric equations.

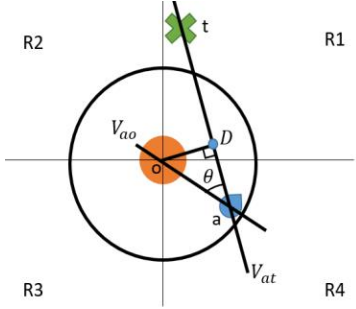

Fig. 1. The world is divided into four regions. 'a', 'o', and 't' are stands for agent, nearest obstacle, and the target. The location and the region of the D-point are depicted here.

First, the vectors from the agent to the target ($\overrightarrow{V_{at}}$) and from the agent to the nearest obstacle ($\overrightarrow{V_{ao}}$) are calculated by

$$\overrightarrow{V_{at}} = (x_t, y_t, z_t) - (x_a, y_a, z_a),$$
$$\overrightarrow{V_{ao}} = (x_o, y_o, z_o) - (x_a, y_a, z_a) \tag{2}$$

where, $(x_o, y_o, z_o)$, $(x_a, y_a, z_a)$, and $(x_t, y_t, z_t)$ are coordinates of the nearest obstacle, the agent, and the target, respectively. Next, we calculate the angel $\theta$ made by $\overrightarrow{V_{at}}$ and $\overrightarrow{V_{ao}}$ vectors, depicted in Fig. 1 by

$$\theta = \cos^{-1}\left(\frac{\overrightarrow{V_{at}} \cdot \overrightarrow{V_{ao}}}{|\overrightarrow{V_{at}}||\overrightarrow{V_{ao}}|}\right), \tag{3}$$

where, $|.|$ is the magnitude and the point in $V_{at}.V_{ao}$ is dot product of the vectors. Then, we define a D-point (Fig. 1) that is the intersection point of two lines, the line which passes through the agent and the target ($V_{at}$), and the line from nearest obstacle orthogonal to the $V_{at}$. Later, to calculate the coordinates of the D-point, we use the CAH (Cosine, Adjacent, Hypotenuse) rule to obtain the distance from agent to the D-point, which is denoted by $d_{Da}$, as

$$d_{Da} = Cos(\theta).|\overrightarrow{V_{ao}}|, \tag{4}$$

where, $\theta$ is the angel between $V_{ao}$ and $V_{at}$. Finally, we calculate the location of the D-point $(x_D, y_D, z_D)$ by travelling the coordinate of the agent in the direction of the target with amount of $d_{Da}$, by

$$(x_D, y_D, z_D) = (x_a, y_a, z_a) + \hat{V}_{at}.d_{Da} \tag{5}$$

where, $\hat{V}_{at}$ is the unit vector, calculated by $\overrightarrow{V_{at}}/|\overrightarrow{V_{at}}|$. Now, to define unique states, we have considered the region of the D-point in the state definition. As an example, suppose the situation shown in Fig. 1. Let's assume that the coordination of the agent, the target, and the obstacle are (4.9, -3.1, 0), (1.2, 13, 0), and (0,0,0), respectively. It's obvious that according to (2), $\overrightarrow{V_{at}} = (-3.7, 16.1, 0)$ and $\overrightarrow{V_{ao}} = (-4.9, 3.1, 0)$. Replacing the values in (3), $\theta$ equals 0.7808 radian. In addition, $d_{Da}$ equals 4.1187 employing (4). Finally, based on (5), the coordination of the D-point is (3.9, 0.91, 0), which is located in R1. In our proposed method the states are define by triples that are the regions belong to the target, the agent and the D-point, respectively. For example, $s = [R1, R4, R1]$ defines the state shown in Fig. 1.

### D. Action Space Definition

Agent's movements in the environment occurs in two areas: safe and non-safe. Suppose that obstacles (physical objects with rigid body) in the environment are surrounded by a circle-shaped area, which is the non-safe area for the agent. The radius of this area is larger than that of the obstacles. In every instance of the time, the agent checks its distance to the nearest obstacle. If the distance is lower than the radius of the non-safe area, it means that the agent is walking in the non-safe area; otherwise, the agent is in the safe area.

According to *Assumption 1*, whenever the agent comes to the safe area, it goes directly toward the target. Thus, when it is in a non-safe area, it should choose a proper action. Options are moving to left or right; the forward action is not considered in order to reduce the number of Q-table entries, and consequently, to reduce the computational complexity. Nevertheless, whenever the forward action is required, a turning right or left action with a small rotation degree is helpful.

The degree of rotation is critical in turning right or left actions. This value depends on the distance from D-point to the nearest obstacle, denoted by $d_{Do}$, which is calculated using Euclidean distance as

$$d_{Do} = \sqrt{(x_D - x_o)^2 + (y_D - y_o)^2 + (z_D - z_o)^2}. \tag{6}$$

In fact, the degree of the rotation has an inverse relation with $d_{Do}$, meaning the less the distance is, the more the degrees of rotation will be. However, according to *Assumption 2*, the degree of rotation cannot be more than 90 degrees. In the training process, the agent can choose the action randomly or it can take the action that causes the value of $d_{Do}$ to be increased.

### E. The Number of States

This section describes the number of states in the states space of the proposed path planning. To calculate the number of states, two different situations should be considered: 1) Target and agent in the same region; and 2) Target and agent in different regions.

When the agent and the target are both located in the same region, the D-point can only be in three regions. For instance, Fig. 2(a) illustrates the situation when both the target and the agent are located in the region R1. As depicted in the figure, based on the location of the agent and the target, the D-point can just be in regions R1, R2, or R4. It never be located in region R3 as the intersection of the two orthogonal line is never located in R3. Furthermore, there are generally four possible ways for the target and the agent to be located in the same region, each one with three possible location of the D-points. Therefore, generally 12 possible number of states exists in which the target and the agent are in the same region ($n_{TASR} = 12$).

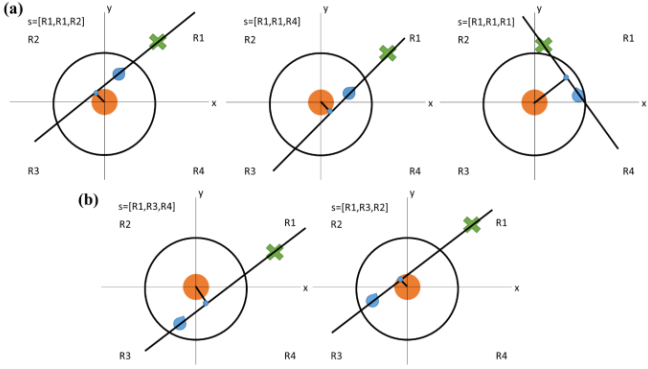

Fig. 2. The possible regions in which the D-point can be located when the target and the agent are in the same region (a) and when they are in different regions (b).

On the other hand, there are 12 different possible combinations, in which the target and the agent are in different regions. In such situations, there are only two possible options for the location of the D-point. We denote the number of states in which the target and the states are in different region by $n_{TADR}$. Hence $n_{TADR} = 24$. Fig. 2(b) depicts an example of such a situation. As can be seen from the figure, when the target and the agent are not in the same region (e.g., R1 and R3), D-point can only be located in two regions. In fact, it never can be located in regions R1 and R3. Finally, the total number of Q table entries is calculated as follows:

$$n_{QTableEntry} = (n_{TASR} + n_{TADR}) * n_a \qquad (7)$$

in which $n_a$ refers to the number of actions which is 2. Thus, the total number of Q table entries, $n_{QTableEntry}$, equals 72.

*F. Immediate Reward Function*

In the reinforcement learning, the way that the operator of the action is punished or rewarded can have a great influence on learning process. As described in Section II.D, obstacles are surrounded by an area called non-safe area. The learning process happens whenever the agent comes to this area. The agent does not get negative reward for entering into this area since the agent does not have any information about the obstacles' locations except the one which is nearest to it. Furthermore, obstacles are moving in a random fashion, which means, in a certain situation, agent can be in a non-safe area and at the same time, another obstacle comes toward it. However, a positive reward is considered for doing the action that leads to take the agent out of the non-safe area (rescue action). In the non-safe area, due to changing $d_{Do}$, we categorize actions into two groups:

1) *involving action*: in which $d_{Do}$ is decreasing; i.e. the D-point is getting close to the obstacle; thus increasing the probability of collision. In this case, the agent is punished with a negative reward.
2) *escape action*: in which $d_{Do}$ is increasing or it does not change. In this case, the agent will receive a neutral reward.

Regarding the above descriptions, the reward $r$ for doing every mentioned actions is defined as follows

$$r = \begin{cases} involving\ action &= -1 \\ escape\ action &= 0 \\ rescue\ action &= 1 \end{cases} \qquad (8)$$

As an example, suppose that the agent is in the situation showed in Fig 1. Let us assume that the radius of the non-safe area is 6.2, the agent's coordination is (4.9, -3.1, 0), and $|\overrightarrow{V_{ao}}|$ equals 5.79. If the agent selects the turning right action, it will receive the rescue reward, i.e. 1, since after doing the action, the value of $|\overrightarrow{V_{ao}}|$ would be greater than 6.2, which means that the agent gets out of the non-safe area.

### III. EXPERIMENTAL RESULTS

*A. Experimental Setup*

For simulations, we have used the Unity3D Game Engine and C# as its programing language. Fig. 3 demonstrates the sequential photos taken from one of the test scenarios showed by the Scene-Tab of Unity3D workspace.

Depending on the evaluation type, scenarios have different setup parameters, which has been presented in Table 1. The fixed parameters are general for overall simulations; but there are some other parameters that their values vary depending on the evaluation type and therefore are considered as variable parameters.

As listed in the first row of Table 1, the frame interval between actions in the non-safe area equals 10 frames in the Unity software. This interval is chosen by trial and error, as the effect of the selected action should be observed in the environment and the agent should receive the corresponding reward. In addition, Table 1 explains that agent movement in each frame is 0.2 in Unity default unit. Therefore, in a single action, agent moves 10*0.2=2 units.

Reported by Table 1, Time threshold for staying in non-safe areas (going from one obstacle's non-safe area to another without any time interval) is 1.5 second. This value is defined by trial and error to detect the trapped situation in which the agent is surrounded by few obstacles. The trapped situation is considered as a failed scenario.

As explained in Table 1, initially the agent, the target and obstacles are located in a 100*100 units area. In random initializing the scene of the simulation, the agent is placed at least five units far from the obstacles to avoid early collisions. Moreover, for a fair evaluation, the setup parameters are same for both our and opponent techniques in all the simulations and experiments.

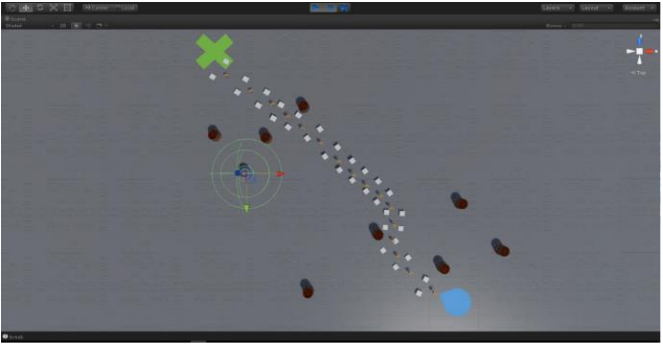

Fig. 3. A view from Unity3D Scene-Tab; sequential photos taken from a test scenario. In this scenario, we stopped the movement of obstacles and target. This figure shows how the agent passes the obstacles in a feasible and short path to reach the goal depicted by cross sign.

### B. The Opponent Approach

To illustrate how our approach improves the path planning in a dynamic environment, we compare the results with the approach presented in [19], as our opponent approach. In the mentioned research, the states are defined by the four regions (R1, R2, R3, and R4); but unlike ours, the regions are related to the nearest obstacle and the target with the robot location as the center of the coordination. To make the states unique from each other, they add the angle made by robot, obstacle, and the target. The range of this angle starts from zero to 360 degrees. Since this interval makes the number of the states too large, they divide it into 8 angular regions (G1 to G8), in which every angular region covers 45 degrees (360 divided by 8). The action space includes three movements called Left, Right, and Forward. The two former are used for obstacle avoidance but the latter is used when the robot is in the safe state.

### C. Evaluation Metrics

First, we compare the performance of our proposed method with the opponent in terms of 1) the number of states; 2) the efficiency of our rewarding approach with opponent, using cumulative received rewards; and 3) the hit rate. Later, we utilize the hit rate to evaluate the effects of changing parameters (e.g., environment density, non-safe area radius, and the number of training scenarios) on the functionality of our approach against the opponent.

#### 1) Number of States

The number of states matters since it has a direct relationship with convergence time, memory overhead, and the problem complexity. So the bigger the number of states (or the number of Q table entries, $n_{QTableEntry}$) is, the more time is required for the convergence.

In the opponent method [19], the states are define by $s = [R_{target}, R_{obstacle}, \theta]$, where $R_{target}$ and $R_{obstacle}$ are the regions of the target and the obstacle, respectively. However, $\theta$ is the angle made by two lines: 1) passing from the target and the agent; and 2) passing from nearest obstacle and the agent. By ab intuitive analysis, according to the state space definitions of the opponent research, the created Q-table has much higher number of entries than ours, i.e., at least twice than ours.

TABLE I
THE DEFINITION OF SCENARIOS' PARAMETERS.

| Parameter Name | value | type |
|---|---|---|
| Frame interval between actions in non-safe area | 10 | Fixed |
| Agent movement in each single frame in non-safe area. | 0.2 (of Unity default unit) | fixed |
| Obstacles and Target movement in each single frame. | From 0.02 to 0.05 (of Unity default unit) | random |
| Time threshold for staying in non-safe areas | 1.5s | fixed |
| The number of obstacles | From 5 to 30 | Variable |
| The number of training scenarios | From 0 to 50 | Variable |
| Initial location of each Obstacles, Target, and Agent | in 100*100 area | random |
| The radius of the non-Safe area | From 3.7 to 6.2 in Unity default Unit | Variable |
| Initial distance between Agent and random Obstacles | Distance > 5 | Variable |
| Initial distance between Agent and Target | 50 <= Distance <= 70 | Variable |
| The degree of agent's rotation | -90 < Degree < 90 | Variable |
| Discount factor ($\gamma$) | 0.9 | fixed |
| Frame interval between actions in non-safe area | 10 | Fixed |
| Agent movement in each single frame in non-safe area. | 0.2 (of Unity default unit) | fixed |
| Obstacles and Target movement in each single frame. | From 0.02 to 0.05 (of Unity default unit) | random |

Compared with $\theta$ that has 8 different cases, the region of D-point in our method, in the worst case, has 3 different options. Thus, the number of states in the opponent method is at least 2.6 (8 divides by 3) times more than ours. The reduction of Q-table's entries leads to faster convergence (more than twice). However, we reduced memory overhead as well as the problem complexity.

#### 2) Cumulative Rewards

The goal of the reward function is to punish the agent by negative rewards when doing wrong actions and giving positive or neutral rewards to show that the chosen actions are proper. The cumulative received reward is an effective measure for comparing reinforcement learning algorithms [25][23]. The higher cumulative reward is, the faster the agent finds effective and suitable actions. Fig. 4 depicts the normalized cumulative received rewards in every 100 rewarding steps for both our and the opponent method. The figure shows the efficiency of the proposed reward function compared with the opponent.

It is worth mentioning that the opponent reward function is based on the distance between robot and the nearest obstacle in a way that after performing the action, if the robot gets closer to the nearest obstacle, then it will receive negative reward but if the robot goes far from the obstacle it receives positive reward. The problem with this definition is that in some cases, although the robot chooses the proper action, it gets closer to the nearest obstacle and receives negative reward.

#### 3) The Hit Rate

The hit rate is the number of successful operations out of 500 testing scenarios. In the successful operations, the agent can reach the goal without colliding with any obstacles. The testing scenario is strongly involved with the action selection in the Q-learning algorithm (Section II.B). Generally, there are two different action selections: exploration and exploitation. In exploration, a random action is chosen for the current state to see which rewards the environment gives. Consequently, it is for finding new uncovered actions or new Q-table entries. On the other hand, in exploitation, the agent benefits from the adapted Q-table. Thus, the algorithm chooses the actions with the maximum Q value in the current state. When referring to the testing scenario, we mainly mean the exploitation phase, in which the Q-table is trained for some epochs.

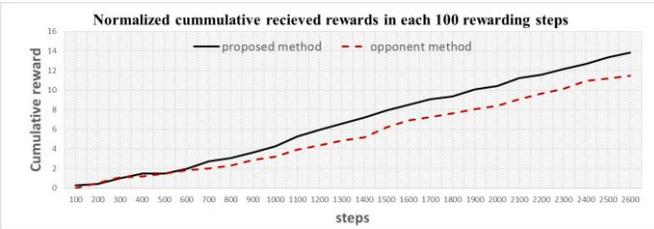

Fig. 4. The comparison of reward functions based on cumulative received rewards in each 100 rewarding steps.

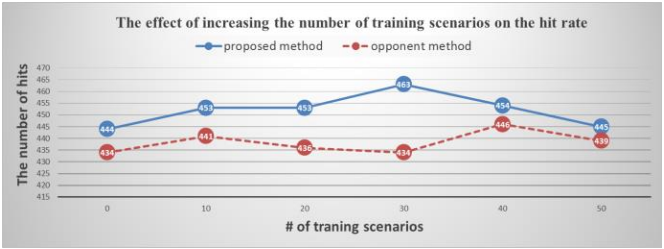

Fig. 5. Comparison of the effect of increasing number of training scenario on the hit rate.

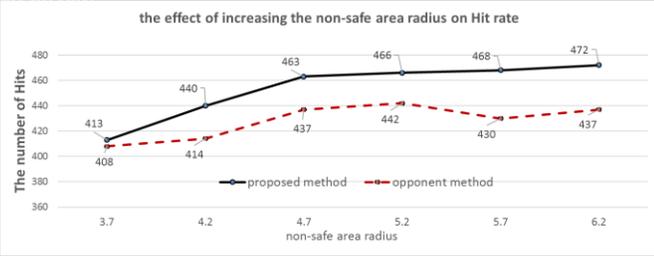

Fig. 6. The effect of changing the radius of the non-safe area on the hit rate.

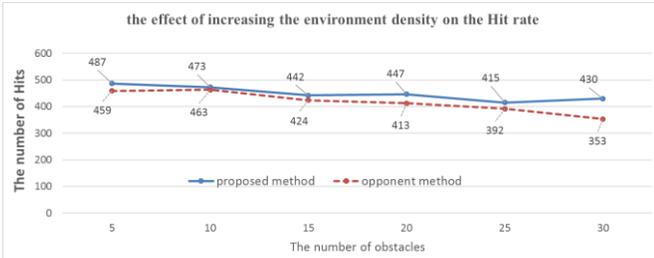

Fig. 7. Increasing the environment's complexity and its effect on the hit rate.

We compare the hit rates in three different experiments to evaluate the effect of the following parameters:
1. the number of training scenarios
2. the radius of the non-safe area
3. the number of obstacles and therefore the complexity of the environment.

In the first evaluation, the effect of increasing the number of training scenarios on the hit rate is analysed. Fig. 5 demonstrates the results of this evaluation as well as the comparison of methods. As this figure illustrates, the proposed method shows two major improvements: 1) reducing the required number of training scenarios, i.e., the agent is trained enough with less effort and less required time; and 2) higher overall hit rate. As can be seen in the figure, even when the number of training scenarios is set to zero, the hit rate is 2% more than the opponent method. Moreover, when comparing the hit rate in the case of choosing the best number of training scenarios of both methods, the hit rate in our method is 3.4% more than the opponent.

As described in Section II, obstacles have a non-safe area. When the agent comes to this area, it starts choosing obstacle avoidance actions. From Fig. 6, it can be inferred that the performance of both methods strongly depends on the radius; although the proposed method shows better performance.

In the next experiment, we have analysed the effect of increasing the environment complexity. As can be seen in Fig. 7, both methods show hit rate reduction when the number of obstacles increases. Furthermore, the proposed method not only improves the hit rate but also show less reduction with increasing number of obstacles. In the opponent method, the difference between hit rates in minimum and maximum number of obstacles is 106. In contrast, when using the proposed method, this number is 57 or in the worst case is 72 which shows that our approach less depends on the environment complexity.

IV. CONCLUSION

In this paper, we introduced a Q-learning method for path planning using a trigonometric method for defining the state space. Using the proposed D-point method, the agents does not need a priori information about the dynamic environment with considering that all the objects in the environment are moving. Because the D-point approach minimized the possible number of states, we decreased the number of entries in the Q-table, which leads to the less memory overhead.

According to our experiments, the D-point method improved the number of successful path planning, and reduced the dependences of a successful operation to the environment parameters. The results illustrated that the D-point approach outperformed the method in [19].

A limitation of our proposed method is that the agent avoids the collision with the nearest obstacle. However, when avoiding the nearest obstacle, another obstacle may come directly toward it. However, if the nearest obstacle moves toward the agent, it does not have enough time to escape. Based on this limitations, the future work will be introducing a new methodology that expand the localization in a way that the agent be cautious of its nearest obstacles and their moving directions. Because the dependency of the proposed approach to the environmental parameters is low, it is applicable not only in the field of video games but also augmented reality applications and soccer robots.